\documentclass[sn-basic]{sn-jnl}

\usepackage{graphicx}%
\usepackage{multirow}%
\usepackage{amsmath,amssymb,amsfonts}%
\usepackage{amsthm}%
\usepackage{mathrsfs}%
\usepackage[title]{appendix}%
\usepackage{xcolor}%
\usepackage{textcomp}%
\usepackage{manyfoot}%
\usepackage{booktabs}%
\usepackage{algorithm}%
\usepackage{algorithmicx}%
\usepackage{algpseudocode}%
\usepackage{listings}%
\usepackage{natbib}

\begin{document}

\title[Article Title]{Zeus: Zero-shot LLM Instruction for Union Segmentation in Multimodal Medical Imaging}


\author[1]{\fnm{Siyuan} \sur{Dai}}
\author[1]{\fnm{Kai} \sur{Ye}}
\author[1]{\fnm{Guodong} \sur{Liu}}
\author*[2]{\fnm{Haoteng} \sur{Tang}}
\author*[1]{\fnm{Liang} \sur{Zhan}}

\affil[1]{\orgdiv{Department of Electrical and Computer Engineering}, \orgname{University of Pittsburgh}, \orgaddress{\street{3700 O’Hara
St}, \city{Pittsburgh}, \postcode{15213}, \state{PA}, \country{USA}}}
\affil[2]{\orgdiv{Department of Computer Science}, \orgname{University of Texas Rio Grande Valley}, \orgaddress{\street{ 1201 W University
Dr}, \city{Edinburg}, \postcode{78539}, \state{TX}, \country{USA}}}


\abstract{Medical image segmentation has achieved remarkable success through the continuous advancement of UNet-based and Transformer-based foundation backbones. 
However, clinical diagnosis in the real world often requires integrating domain knowledge, especially textual information. Conducting multimodal learning involves visual and text modalities shown as a solution, but collecting paired vision-language datasets is expensive and time-consuming, posing significant challenges.
Inspired by the superior ability in numerous cross-modal tasks for Large Language Models (LLMs), we proposed a novel Vision-LLM union framework to address the issues. Specifically, we introduce frozen LLMs for zero-shot instruction generation based on corresponding medical images, imitating the radiology scanning and report generation process.
{To better approximate real-world diagnostic processes}, we generate more precise text instruction from multimodal radiology images (e.g., T1-w or T2-w MRI and CT). Based on the impressive ability of semantic understanding and rich knowledge of LLMs. This process emphasizes extracting special features from different modalities and reunion the information for the ultimate clinical diagnostic.
With generated text instruction, our proposed union segmentation framework can handle multimodal segmentation without prior collected vision-language datasets.
To evaluate our proposed method, we conduct comprehensive experiments with influential baselines, the statistical results and the visualized case study demonstrate the superiority of our novel method.}

\keywords{Union Segmentation, Multimodal Learning, Large Language Model, Instruction Prompt.}


\maketitle

\section{Introduction}
Medical imaging analysis is pivotal for analyzing radiology information, targeting areas significant for clinical diagnosis \cite{azad2022medical,dai2024constrained,minaee2021image}, and biomedical research \cite{ye2025bpen,tang2024instantaneous}. 
The rise of foundation models and expansive medical image datasets has revolutionized this domain, offering precise and efficient automated segmentation. 
Such progress aids in-depth medical imaging research, thereby increasing the accuracy of clinical diagnoses and treatments.

Despite the advancements in 2D and 3D medical image segmentation \cite{wang2022medical,wang2022medclip,tang2024ex}, these researches mostly focus on a single medical modality (e.g., T1-w or T2-w MRI and CT). They tried to train and test the model on the same modality and hardly ever involved scanning the same part of the body under different modalities. Some competitions \cite{menze2014multimodal,kavur2021chaos,antonelli2022medical} collected some datasets in such a setting, but they still explored the performance of one input and one output, mainly concentrated on domain shift problem or transfer learning.
Nevertheless, the intricacies of multimodal medical imaging \cite{zhang2022mmformer} recognize the superior capabilities of scans from multiple modalities over single-modality imaging. 
These modalities provide distinct and complementary insights into tissue anatomy, functionality, and pathology. Furthermore, physicians always diagnose based on multimodal radiology image data. They have multiple input images and analyses of the same organs or lesions for one diagnosis output. 

Meanwhile, only combining previous foundation models for multimodal medical image segmentation is still naive. They did have achieved great progress in medical image segmentation \cite{guo2019deep,zhao2022modality,zhang2022mmformer,dai2025sin}, however, these models overlook cross-domain knowledge, such as textual information, which could be regarded as the medical knowledge in the textbook when training a physician. They easily regarded them as augmented data for neural network training without highlighting the unique information each modality brings.
Medical training emphasizes radiological image explanation and understanding combined with the reasoning of text-based general medical knowledge, which further highlights a gap in current techniques. 
To address such an issue, and align the diagnosis process in real life, we highlight the importance of text-based domain knowledge as modality-related information for assisting medical image analysis, proposing a novel vision-language union framework based on a novel powerful multimodal segmentation backbone \cite{kirillov2023segment}. Our work is conducted in a special situation combining cross-modal knowledge, to distinguish multimodal learning in the general domain (vision, language, audio, etc) with multimodal image segmentation in the biomedical domain, we regard such a situation as Union Segmentation and specific reference to multimodal segmentation as multimodal learning.

Furthermore, collecting paired cross-modal (vision and language) datasets is expensive and time-consuming. Especially when it comes to the medical domain which with strict privacy restrictions. Recently, Large language models (LLMs) have extended their impact beyond text-only applications, showing proficiency in various domains such as game, vision, and FPGA \cite{cui2024survey,light2023text,zhu2023minigpt,fu2023gpt4aigchip}. With such a pivotal advancement, LLMs have been bridging the gap across different modalities, especially between the vision and language domains.
For example, LISA \cite{lai2023lisa} and GLaMM \cite{rasheed2023glamm} are notable for integrating LLMs into pure vision tasks, they expanded the original text-based vocabulary by introducing a new token \textbf{$<$ SEG $>$} to push the request for binary segmentation output. Although they also freeze the entire LLMs, they needed to pre-train the additional MLP layers and the LoRA\cite{hu2021lora} under extra huge datasets and fine-tune the model for their own elaborated datasets then for down-stream tasks, which is computationally intensive, time-consuming, and not an end-to-end framework.
To solve the challenges above, we propose a novel method for zero-shot instruction generation based on a frozen LLM, and such a method does not need additional datasets for pre-training and fine-tuning, constructing an end-to-end union segmentation framework. Our proposed framework is also a promising exploration of the zero-shot ability of LLMs to dig domain knowledge.

To this end, we introduce Zeus, an end-to-end union segmentation framework designed by a powerful multimodal segmentation backbone \cite{kirillov2023segment} and guided by a pretrained large vision language model (LVLM) \cite{wang2022medclip} with a pretrained LLM \cite{chiang2023vicuna} for multimodal medical imaging without extra pre-training or fine-tuning. 
We evaluate our proposed framework on three publicly available multimodal datasets, including the MSD-Prostate, MSD-Brain \cite{antonelli2022medical}, and CHAOS \cite{kavur2021chaos}. 
Our main contribution to this article is summarized below:
\begin{itemize}
\item{We introduce a novel end-to-end union segmentation framework for bridging the current multimodal medical image segmentation task and the clinical diagnosis process in real life, imitating the physician for considering radiology images from multiple modalities with their extensive domain knowledge.}
\item{We introduce LLMs and LVLMs for generating text instruction for digging domain knowledge, exploring the zero-shot ability to understand semantic features in a cross-modal situation.}
\item{We conduct extensive experiments on 3 public datasets and compare them with influential end-to-end baselines under the three different multimodal learning settings, showing the superiority of our proposed method and the promising ability of LLMs.}
\end{itemize}

\section{Related work}
\subsection{Multimodal Learning}
Multimodal learning is a promising paradigm to integrate data from various sources to improve decision-making and predictions and has seen significant advancements \cite{ngiam2011multimodal,baltruvsaitis2018multimodal,xu2023multimodal,yin2024heterogeneous}. Features from different views (e.g. visual, text, audio, etc.) can provide more comprehensive representation information for semantic understanding. It is hard to continuously improve the effectiveness of representation learning with a single modality. Pre-trained vision-language model \cite{lu2019vilbert,li2019visualbert} significantly improves the performance both in the vision and language tasks after doing multimodal learning from combined sources. Besides, multimodal learning shows great importance in autonomous driving \cite{xiao2020multimodal}, generative model \cite{suzuki2022survey}, healthcare \cite{muhammad2021comprehensive,ye2023bidirectional,tang2024interpretable}. However, in the field of medical imaging diagnosis, modality is a much more fine-grained concept than multi-source data such as image, audio, text, etc., and different modalities can exist for the same object (e.g. organ, lesion, etc.). This concept is particularly valuable in medical imaging, where different imaging modalities (e.g., MRI-T1, MRI-T2, CT, X-ray, etc) can provide complementary information about the same anatomical structures from different views. \cite{dalmaz2022resvit,zhang2022mmformer,guo2019deep} tried to fuse the images from different modalities and help the medical image analysis process, but they never highlighted specific information about different modalities and they used the different modules to process different modalities which are computationally expensive and time-consuming. {LViT \cite{li2023lvit} explored annotating medical images with additional text labels to assist lesion segmentation. However, the text information they utilized is specific to lesion segmentation and not intended for medical diagnosis purposes.} Above all, there isn't a benchmark for considering such a significant medical image problem.

\subsection{Large Language Model-based Vision Language Model}
Recently, LLMs \cite{achiam2023gpt,touvron2023llama} have extended their impact beyond text-related applications, including multi-agent, chip design, coding, etc. for showing promising ability in conversation, reasoning, and planning, etc. \cite{cui2024survey,light2023text,zhu2023minigpt,fu2023gpt4aigchip}. With such a pivotal advancement, LLMs have been bridging the gap across different modalities, especially between the vision and language domains.
More than GPT family \cite{achiam2023gpt}, Flamingo \cite{alayrac2022flamingo}, BLIP-2 \cite{li2023blip}, LLAVA \cite{liu2024visual} also establish a connection between visual perception and human languages, showcasing impressive in-context few-shot learning capabilities for visual semantic understanding and reasoning.
Meanwhile, LISA \cite{lai2023lisa}, VisionLLM \cite{wang2024visionllm} and GLaMM \cite{rasheed2023glamm} are notable for using LLMs in vision-centric tasks. However, previous works always require fine-tuning LLM for their specific datasets, even modifying the vocabulary, which is computationally intensive and time-consuming.
When it comes to the medical domain, researchers \cite{thawkar2023xraygpt,wang2022medclip} utilized the fine-tuned LLMs and LVLMs and trained on a vast collection of medical-related image-text pairs \cite{johnson2019mimic,demner2016preparing}. Nevertheless, the exploration in the medical vision domain still focuses on easy captioning tasks.

\subsection{Pre-trained Large Language Model In Medical Domain}
LLMs in the medical domain start from the pure text-based tasks for biomedical research and medical question-answer for patients \cite{singhal2023large}. However, when generating a long context, a huge knowledge gap exists between most of the medical LLMs and the real doctors. Instructing the mechanisms like instruction-prompting, chain-of-thought, etc., things would be better \cite{singhal2023towards,li2023chatdoctor}, so does the bilingual scenario \cite{wang2023huatuo} and biomedical science \cite{taylor2022galactica}. More complex tasks need more comprehensive and huge datasets with novel algorithms. Vision-centric multimodal tasks in the general domain own near-infinite web images and captions e.g., Flickr \cite{joulin2016learning} and COCO Captions \cite{desai2021virtex}, which dwarfs the scale of medical image-text data. paired images and captions from the general domain. Likewise, existing methods in the general domain make it hard to align the cross-modal retrieval and hence do not support zero-shot predictions for applying to the medical domain. After the success based on the open-source of LLMs \cite{touvron2023llama,chiang2023vicuna,taori2023stanford}, researchers could take advantage of the pre-trained LLMs on the general domain and fine-tuned on smaller medical datasets \cite{thawkar2023xraygpt,wang2022medclip,li2023chatdoctor}. Previous works show great advancements in encapsulating the semantic understanding ability of LLMs, involving the planning and subject localization for vision-language tasks. 
Current applications of LLMs in medical images primarily target one radiology image captioning, however, the area of multimodal medical image segmentation remains largely explored and it is closer to real-world clinical application.

\section{Methodology}
Our proposed Zeus aims to imitate a real-world physician to make a diagnosis, combining multimodal images with corresponding text instructions for union segmentation. Due to our goal of exploring the zero-shot ability of pre-trained LLMs and LVLMs, we adopt the vision encoder from MedCLIP \cite{wang2022medclip} as our vision encoder and the Vicuna \cite{chiang2023vicuna} as our used LLM as the first part of our Zeus framework to analyze the multimodal images and generate instruction prompts, which encapsulate modality descriptions. It is worth mentioning that we use the Vicuna-Rad weight from XrayGPT \cite{thawkar2023xraygpt} in order to make the knowledge spaces for image and language well aligned. For final mask prediction, we then employ the mask decoder from SAM, which could conduct union segmentation with image and text instruction as two inputs.

\subsection{Framework of Zeus}
\begin{figure*}[t]
\includegraphics[width=\textwidth]{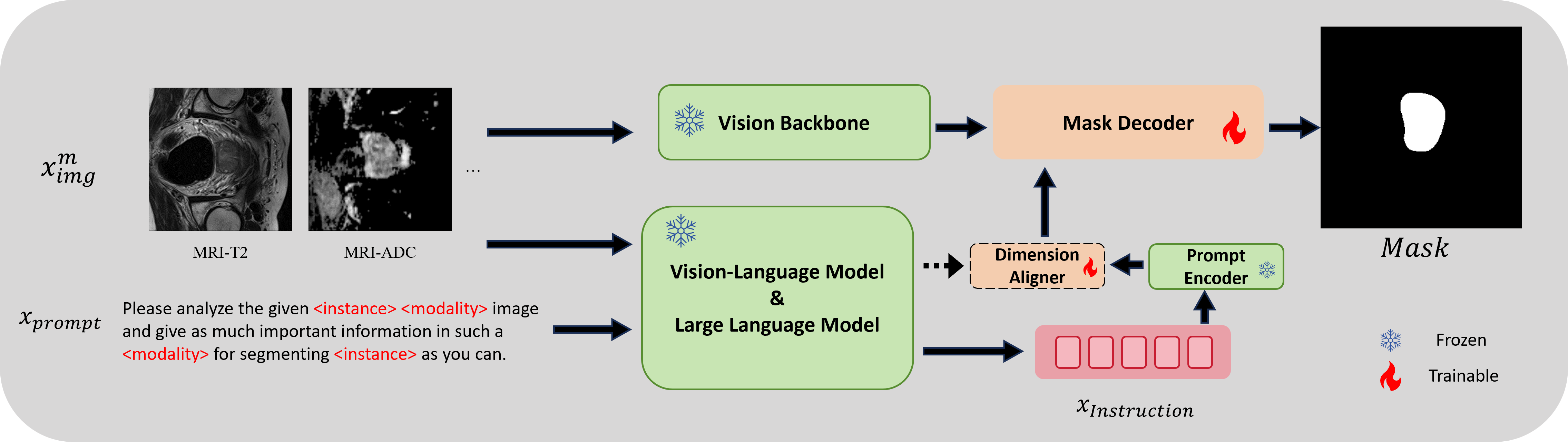}
\caption{The architecture of Zeus. It consists of a pre-trained vision-language model and a large language model with a pre-trained vision backbone as the prompt encoder. The trainable mask decoder accepts image-instruction pairs for mask prediction.} \label{fig:frame}
\end{figure*}
We show in Fig.\ref{fig:frame} an overview of our Zeus. Conventional segmentation models typically employ a U-Net encoder \cite{ronneberger2015u} or integrate Transformer blocks \cite{chen2021transunet} for image encoding, feature extraction, and down-sampling. However, these decoders are not well-suited for simultaneously processing image embeddings and text instructions. In this regard, we adopt the SAM \cite{kirillov2023segment} as our mask predictor $F_{pred}$, designed specifically for semantic segmentation with various types of prompts (e.g., text, points, bounding boxes, etc.). While other flexible options exist \cite{cheng2022masked}. In order to align our encoder $F_{enc}$ with the mask predictor $F_{pred}$ and take great advantage of pre-trained ability from SAM, we adopt the core configuration and parameter settings used in SAM \cite{kirillov2023segment} to ensure compatibility. {For the Instruction generation part, we first input the images into the MedCLIP \cite{wang2022medclip} vision encoder and then combine them with a prompt before sending them to the LLM backbone. The instruction generated by the LLM is then processed by a prompt encoder, also from MedCLIP. Finally, vision embedding and dimension-aligned prompt embedding are used for union segmentation by the SAM decoder. The modules are listed in Table \ref{table: modules}.}

\begin{table}[t]
    \centering
    \caption{{Detailed information about the used modules.}}
    \begin{tabular}{lcc} \hline
    {Module} & {Symbol} & {Trainable}\\ \hline
    {SAM vision encoder} & {$F_{enc}$} & {$\times$} \\ 
    {VLM vision encoder} & {$\Tilde{F_v}$} & {$\times$} \\ 
    {Projection layer between VLM and LLM} & {$f_{vt}^{\theta}$} & {$\surd$} \\
    {LLM backbone} & {$F_{LLM}$} & {$\times$} \\
    {Instruction prompt encoder} & {$\Tilde{F_t}$} & {$\times$} \\
    {Dimension Aligner} & {$f_{tt}^{\theta}$} & {$\surd$} \\
    {Mask decoder} & {$F_{pred}^{\theta}$} & {$\surd$} \\ \hline
    \end{tabular}
    \label{table: modules}
\end{table}

Given an image input set $x_{img}$, each set contains samples in different modalities $\{x_{img}^{m}= (x_{img}^{1}, ..., x_{img}^{M})\}$, where $M$ represents the number of modalities for the corresponding image. Typically, the input image resolution is $1024 \times 1024$ which is aligned with SAM. After processing by the encoder $F_{enc}$, the embedding of the encoded image is represented in the format \textbf{$C\times H \times W$}, indicating that any conventional image encoder backbone can be employed. Following the setting with SAM, the output image embedding $V_e$ has a resolution of $64\times 64$, resulting in a $16\times$ downscaling of the input image. It is worth mentioning that the whole encoder $F_{enc}$ is fully frozen. 

\begin{equation}
  V_e = F_{enc}(x_{img})  
\label{eq:vision_embedding}
\end{equation}

For the text instruction generation, each given $x_{img}^{m}$ is associated with a text prompt $x_{prompt}^{m}:$ 
\textit{"Please analyze the given $\langle$instance$\rangle$ $\langle$modality$\rangle$ image and give as much important information in such a $\langle$modality$\rangle$ for segmenting $\langle$instance$\rangle$ as you can."}
and the $\langle$instance$\rangle$ is replaced with the target name (e.g. organs, tissues), and the $\langle$modality$\rangle$ is substituted with the modality of the input image (e.g. MRI-T2, MRI-ADC). 
The specific text instruction paired with each image is generated by a large vision-language model (LVLM) and a large language model (LLM). The default prompt format is textual, but it can also be the last-layer embedding if an open-source LLM (e.g., LLaMA) is utilized. For the given $x_{img}^{m}$, it is processed the second time for the instruction generation, this vision encoder $\Tilde{F}_{v}$ shares the same design with $F_{enc}$, a pre-trained ViT, but the encoder is tuned by MedCLIP \cite{wang2022medclip} for aligning the vision and language knowledge into the medical domain, and the input resolution is $256\times 256$. Even though MedCLIP already added an extra projection head after a ViT encoder which is used for downstream tasks, we froze all the modules from MedCLIP and used an additional trainable two-layer MLP projection before processing by LLMs.
\begin{equation}
\begin{aligned}
  \Tilde{V}_e & = \Tilde{F}_{v}(x_{img}) \\
  x_{instruct} & = F_{LLM}(f_{vt}^{\theta}(\Tilde{V}_e), x_{prompt}^{m})
\end{aligned}
\label{eq:vision_instruct}
\end{equation}
Where $\Tilde{F}_{v}$ is the vision encoder of MedCLIP \cite{wang2022medclip} and $f_{vt}^{\theta}$ is the followed extra projection head. The expected well-aligned image embedding $\Tilde{V}_e$ could be regarded as a special language, processing by the LLM $F_{LLM}$ with its corresponding prompt $x_{prompt}^{m}$. $F_{LLM}$ is a frozen Vicuna model and the weight is pre-trained from Vicuna-Rad.

When the image embedding $V_e$ and paired instruction $x_{instruction}$ are obtained, $F_{pred}$ should accept two inputs. {Following the setup of SAM \cite{kirillov2023segment}, the dimension of the image embedding $V_e$ matches that of the decoder input. As a result, no additional modules are required between $F_{enc}$ and $F_{pred}^{\theta}$.} As our defaulted instruction is in text format, an additional instruction prompt encoder $F_{p\_enc}$ generates the instruction embedding. 
We aim to follow the instruction prompt encoder in \cite{kirillov2023segment} for text instruction, as it is trained in conjunction with the $F_{enc}$ and $F_{pred}$, allowing us to utilize the pre-trained model. 
However, the publicly released SAM code does not include a text-based prompt segmentation procedure. Therefore, for another alignment between the image and the text embeddings, we adopt the text encoder from MedCLIP \cite{wang2022medclip} as our instruction prompt encoder $F_{p_{enc}}$, which is well-aligned with the image encoder $F_{enc}$ in previous instruction generation module. We use pre-trained checkpoints to leverage its natural alignment between medical text and vision. Additionally, we aim to employ our text instruction as the only sparse prompt in SAM, replacing other sparse prompts such as points and bounding boxes that are naturally employed in original SAM. The dimension of these sparse prompts is $256$, while the default dimension of the text embedding from MedCLIP is $512$. To address this discrepancy, we introduce another linear projection $f_{tt}^{\theta}$ to map the dimensions accordingly. It is worth mentioning that before our trainable projection layer, there was a linear layer after the text encoder in MedCLIP, we also kept such a setting and froze it. 
The overall process can be formulated as Eq.\ref{eq:frame1}
\begin{equation}
\begin{aligned}
  & e_{instruct}^m = \Tilde{F}_{t}(x_{instruct}^m) \\
  & Mask = F_{pred}^{\theta }(V_{e}^m, f_{tt}^{\theta}(e_{instruct}^m)) \\
\end{aligned}
\label{eq:frame1}
\end{equation}
Where $\theta$ represents the trained parameters of each associated module, $V_{e}^m$ and $e_{instruct}^m$ denote the embeddings of the image, and the text instruction, respectively, and $x_{instruct}^m$ is the output text format instruction. $\Tilde{F}_{t}$ is the text encoder from MedCLIP \cite{wang2022medclip}.

In cases where the instruction is already in an embedding format, the two-layer projection layer $f_{tt}^{\theta}$ is also applied to align the dimension of the raw text embedding with that of the input sparse prompt embedding.

For a typical text embedding, the shape format is \textbf{$L \times H$}, where $L$ represents the length of the text and $H$ is the dimension of each text. In our configuration, we treat the entire generated instruction as a single sentence and perform encoding without additional preprocessing or split, resulting in a processed text embedding shape of $1 \times 256$.
Our mask decoder performs after several text-vision alignment processes with the paired image and text embeddings.
Initially, text embeddings undergo conventional self-attention. Subsequently, cross-attention from text embedding to image embedding is implemented for prior alignment, followed by reverse cross-attention from image to text after a one-layer MLP projection for post-cross-modal knowledge alignment. This procedure is repeated twice.
After extracting cross-modal information, the updated embedding is upsampled by two transposed convolutional layers, resulting in a shape four times larger than the original embedding, with both kernel size and stride equal to $2$. This is distinct from conventional decoders that upscale using interpolation methods and a $3 \times 3$ kernel with stride and padding equal to $1$.
The shape is now four times smaller than the desired output mask size. 
Another cross-modal alignment is performed using a small 3-layer MLP with cross-attention from text to image. This attention is applied to perform a spatial point-wise product with the upsampled image embedding to enhance cross-modal knowledge. In contrast to the original SAM model, we resize the mask to $256 \times 256$ and compute the loss and metrics directly, without estimating the $IoU$ score in the middle of the network. {If additional downstream vision-centric tasks are to be explored, extra MLP projection layers (e.g., a classification head or an object detection head) can be introduced after the decoder to align dimensions and modify the loss function for specific purposes.}

\begin{figure*}[t]
\includegraphics[width=\textwidth]{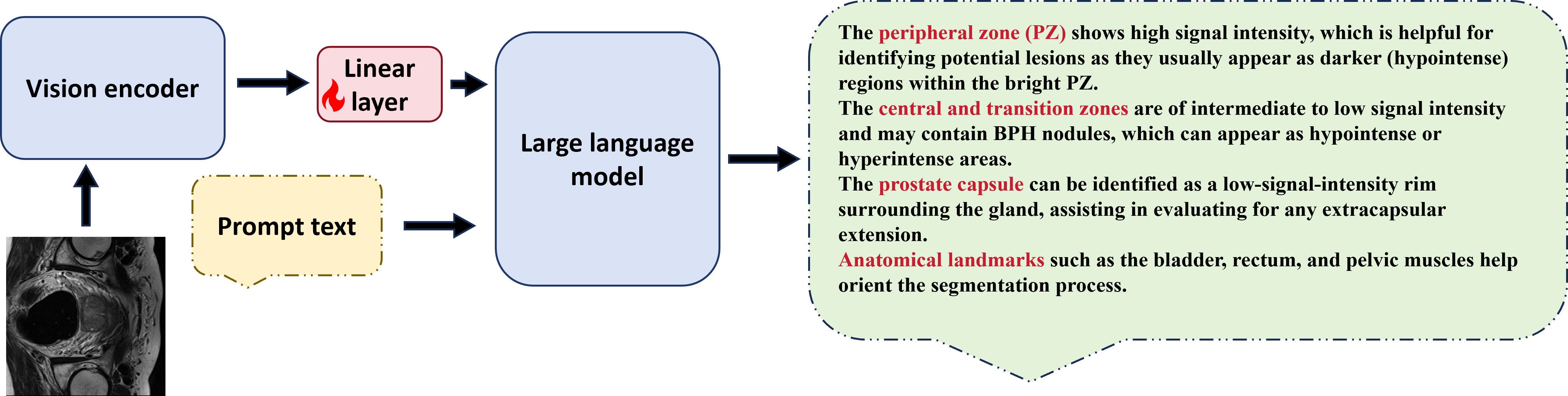}
\caption{The pipeline of text instruction generation. It consists of a pre-trained vision-language model but only processes the image. A simple linear layer is used for knowledge transformation from a vision-language model to a pure large language model with text prompts.} \label{fig:vlm}
\end{figure*}

\subsection{Zero-shot Instruction Generation}
Traditional LLMs cannot directly process image embeddings without the addition of special tokens and extensive fine-tuning, which is computationally expensive and necessitates a large training dataset. As shown in Fig. \ref{fig:vlm}, we aim to utilize the current pre-trained model for zero-shot instruction generation without requiring additional datasets, making conventional pure vision-based encoders \cite{he2016deep,han2022survey} unsuitable for our purposes. We adopt the MedCLIP \cite{wang2022medclip} architecture as our vision encoder, which is specifically designed for image captioning and vision-language alignment. Additionally, other backbones designed for vision-text alignment purposes \cite{wang2022medclip} are also viable options for our framework.

We employ Vicuna-Radiology \cite{thawkar2023xraygpt,chiang2023vicuna} as the employed LLM, which is fine-tuned on extensive image-text paired medical datasets, building upon the original LLAMA \cite{touvron2023llama} model checkpoint. Given that MedCLIP and Vicuna models are not originally trained on identical datasets, the two modules also needed to be frozen and aligned with our used datasets, we facilitate their connection through a trainable two-layer linear projection for projecting the image-level features, represented by $\Tilde{V}_e$, into corresponding language embedding tokens. The pre-trained model owned two text queries in the overall LLM module followed \cite{thawkar2023xraygpt}. The first query, denoted as \textit{\#\#\#Assistant}, serves the purpose of determining the system role, which is defined as "You are a helpful healthcare virtual assistant." when training by \cite{thawkar2023xraygpt}. The second text query, \textit{\#\#\#Doctor}, corresponds to the instruction prompt.
For our zero-shot generating, the LLM incorporates an internal prompt:
\textit{\#\#\#Doctor: $X_RX_Q$} \textit{\#\#\#Assistant: $X_S$}

In this context, $X_R$ will be replaced by the image embedding generated by the MLP layer, $X_Q$ represents the text prompt $x_{instruct}$ that we input, and $X_S$ is the output text instruction $x_{instrut}$. If the text is available, the last-layer embedding of $e_{instruct}$ could be directly connected with $f_{tt}^{\theta}$ for mask prediction.
In our case, to align the vision and text modality by MedCLIP \cite{wang2022medclip}, we denote the LLM with the combined tokenizer and text de-tokenizer for the input and output format are all in text.
\begin{equation}
\begin{aligned}
  e_{instruct} & = F_{LLM}^{t}(f_{vt}^{\theta}(\Tilde{V}_e))\\
  x_{instruct} & = F_{LLM}^{dt}(e_{insttuct})
\end{aligned}
\label{eq:llm}
\end{equation}
Where $F_{LLM}^{t}$ is the tokenizer of the LLM and $F_{LLM}^{dt}$ represents the de-tokenizer to generating text instruction.

\subsection{Problem Definition and Benchmark}
Since most of the current influential medical image segmentation models only focused on a single modality. Given a set of medical images $x^M$ for one subject under multimodal imaging and output a comprehensive result for the target subject, sharing a similar formulation with the previous segmentation task \cite{guo2019deep}. To consider other conventional influential baselines, we give three different scenarios for this benchmark: 1) early fusion for merging the multimodal images at the beginning of the input at the channel-wise dimension; 2) hybrid fusion for merging at the representation level which means that different image modality will have different encoders but shared the same decoder; 3) late fusion for different modality under different whole frameworks and combine the mask at the end.

\subsection{Baseline and Evaluation Metrics}
Under the US benchmark, every framework needed to be the end-to-end architecture to fit the three fusion strategies. So we only involve end-to-end baselines and also do not consider cascade models like \cite{isensee2021nnu,li2018h}, otherwise the parameters will become way more huge and it would have too much constriction in clinical applications. We choose seven influential segmentation models to validate our proposed Zeus framework.
These included UNet++ \cite{zhou2019unet++}, AttUNet \cite{oktay2018attention}, ResUNet \cite{xiao2018weighted}, UNeXt \cite{valanarasu2022unext}, the original U-Net \cite{ronneberger2015u}, TransUNet \cite{chen2021transunet}, and MISSFormer \cite{huang2022missformer}.
The original U-Net and TransUNet were recognized for their robust performance in medical image segmentation.
UNet++, AttUNet, Res-UNet, and UNeXt have been influential U-Net-based models in this domain, while TransUNet and MISSFormer are renowned for their transformer-based architectures. 
Performance evaluation was conducted using {two} metrics: mean intersection over union (mIoU), and Dice similarity coefficient (DSC).
mIoU and DSC are overlap-based metrics, we express them as percentages, with higher values signifying superior performance.

\begin{table}[t]
\centering
\caption{Quantitative results of different methods on three datasets. The best results are shown in bolded font and the second best is underlined. *: Our proposed Zeus is not applicable for early fusion and we only use a single decoder for mask prediction but still processed different modalities one by one for the hybrid fusion.}
\setlength{\tabcolsep}{1.5mm}
{
\begin{tabular}{ccccccccc}
\hline
\multirow{2}*{Networks} &\multirow{2}*{Fusion} & \multicolumn{2}{c}{CHAOS} & \multicolumn{2}{c}{MSD-Prostate} & \multicolumn{2}{c}{MSD-Brain} & \multirow{2}*{Params$\downarrow$}\\ 
 & & \multicolumn{1}{c}{DSC$\uparrow$}  &  mIoU$\uparrow$ & \multicolumn{1}{c}{DSC$\uparrow$} & mIoU$\uparrow$ & \multicolumn{1}{c}{DSC$\uparrow$} & mIoU$\uparrow$ \\ \hline 
\multirow{3}*{U-Net} & early & \multicolumn{1}{c}{79.25} & {78.31} & \multicolumn{1}{c}{63.64} & {63.20} & \multicolumn{1}{c}{74.10} & {73.08} & 18.31M\\ 
                    & hybrid & \multicolumn{1}{c}{80.05} & {78.66} & \multicolumn{1}{c}{65.89} & {63.91} & \multicolumn{1}{c}{75.99} & {73.16} &40.04M \\
                    & late & \multicolumn{1}{c}{81.76} & {\underline{80.50}} & \multicolumn{1}{c}{67.30} & {65.88} & \multicolumn{1}{c}{76.22} & {\underline{75.31}} & 54.94M \\\hline
\multirow{3}*{AttUNet} & early & \multicolumn{1}{c}{79.78} & {78.19} & \multicolumn{1}{c}{62.18} & {64.22} & \multicolumn{1}{c}{72.01} & {70.20} & 27.06M\\ 
                    & hybrid & \multicolumn{1}{c}{78.96} & {77.31} & \multicolumn{1}{c}{63.41} & {63.21} & \multicolumn{1}{c}{72.30} & {70.67} & 56.38M \\
                    & late & \multicolumn{1}{c}{79.11} & {77.39} & \multicolumn{1}{c}{65.02} & {64.13} & \multicolumn{1}{c}{74.26} & {72.97} & 94.08M \\\hline
\multirow{3}*{ResUNet} & early & \multicolumn{1}{c}{78.21} & {76.09} & \multicolumn{1}{c}{64.70} & {63.19} & \multicolumn{1}{c}{73.22} & {72.69} & 34.06M\\ 
                    & hybrid & \multicolumn{1}{c}{79.35} & {74.16} & \multicolumn{1}{c}{65.86} & {63.10} & \multicolumn{1}{c}{74.31} & {70.66} &70.35M \\
                    & late & \multicolumn{1}{c}{81.33} & {79.96} & \multicolumn{1}{c}{65.24} & {61.11} & \multicolumn{1}{c}{74.55} & {71.29} & 110.82M \\\hline
\multirow{3}*{UNeXt} & early & \multicolumn{1}{c}{72.91} & {70.02} & \multicolumn{1}{c}{60.10} & {58.33} & \multicolumn{1}{c}{70.22} & {68.31} & 5.04M\\ 
                    & hybrid & \multicolumn{1}{c}{74.31} & {70.68} & \multicolumn{1}{c}{64.61} & {63.20} & \multicolumn{1}{c}{72.38} & {70.91} & 12.86M\\
                    & late & \multicolumn{1}{c}{80.62} & {77.28} & \multicolumn{1}{c}{68.15} & {66.21} & \multicolumn{1}{c}{75.30} & {72.27} & 18.66M\\\hline
\multirow{3}*{UNet++} & early & \multicolumn{1}{c}{77.53} & {75.09} & \multicolumn{1}{c}{61.42} & {59.18} & \multicolumn{1}{c}{73.89} & {72.02} & 20.03M\\ 
                    & hybrid & \multicolumn{1}{c}{79.01} & {76.83} & \multicolumn{1}{c}{65.66} & {62.37} & \multicolumn{1}{c}{76.22} & {73.20} &42.16M \\
                    & late & \multicolumn{1}{c}{80.60} & {77.66} & \multicolumn{1}{c}{65.17} & {61.55} & \multicolumn{1}{c}{76.30} & {71.54} & 64.48M \\\hline
\multirow{3}*{TransUNet} & early & \multicolumn{1}{c}{78.19} & {76.26} & \multicolumn{1}{c}{65.14} & {64.03} & \multicolumn{1}{c}{74.19} & {73.15} & 112.40M\\ 
                    & hybrid & \multicolumn{1}{c}{79.22} & {76.99} & \multicolumn{1}{c}{67.36} & {64.11} & \multicolumn{1}{c}{75.30} & {72.21} & 308.26M\\
                    & late & \multicolumn{1}{c}{\underline{82.31}} & {\underline{80.09}} & \multicolumn{1}{c}{68.10} & {\underline{66.72}} & \multicolumn{1}{c}{76.13} & {74.89} & 315.96M\\\hline
\multirow{3}*{MISSFormer} & early & \multicolumn{1}{c}{78.66} & {75.31} & \multicolumn{1}{c}{64.02} & {62.85} & \multicolumn{1}{c}{71.90} & {70.11} & 188.08M\\ 
                    & hybrid & \multicolumn{1}{c}{76.72} & {73.45} & \multicolumn{1}{c}{66.46} & {64.12} & \multicolumn{1}{c}{75.15} & {72.96} & 383.47M\\
                    & late & \multicolumn{1}{c}{81.16} & {78.45} & \multicolumn{1}{c}{\underline{69.59}} & {\underline{66.81}} & \multicolumn{1}{c}{77.55} & {74.18} & 610.88\\\hline
\multirow{2}*{Zeus*} & hybrid & \multicolumn{1}{c}{\underline{82.70}} & {79.07} & \multicolumn{1}{c}{66.30} & {63.32} & \multicolumn{1}{c}{\underline{78.30}} & {\underline{75.18}} & \textbf{8.06M}\\
                    & late & \multicolumn{1}{c}{\textbf{85.80}} & {\textbf{84.19}} & \multicolumn{1}{c}{\textbf{71.09}} & {\textbf{68.36}} & \multicolumn{1}{c}{\textbf{83.67}} & {\textbf{80.86}} & \underline{12.44M} \\\hline
\end{tabular}
}
\label{table:com}
\end{table}

\subsection{Datasets}
Three publicly available datasets were used to evaluate our framework, including the MSD-Prostate, the MSD-Brain \cite{antonelli2022medical}, and the abdominal organ segmentation (CHAOS) \cite{kavur2021chaos}. The first two datasets are from the MSD challenge, aimed at advancing medical image segmentation.
\begin{itemize}
\item{\textbf{MSD-Brain}: The MSD-Brain dataset is a part of the Medical Segmentation Decathlon (MSD). It is a comprehensive dataset designed to facilitate research and development in brain tumor segmentation, which encompasses a variety of MRI sequences such as T1, T1-Gd (T1 with gadolinium contrast), T2, and FLAIR (Fluid-Attenuated Inversion Recovery). It provides $484$ 3D multimodal volumes with pixel-level annotations for different brain tumor structures, including the whole tumor, tumor core, and enhancing tumor regions. We only do bi-class segmentation for our experiments, all the annotated regions will be mapped to be tumors.}
\item{\textbf{MSD-Prostate}: The MSD-Prostate dataset is also a part of the MSD, which is a specialized dataset aimed at advancing the field of prostate cancer. This dataset is instrumental for developing and evaluating algorithms designed to segment prostate structures in multimodal medical imaging. It provides $32$ 3D multimodal MRI scans by the sequence of T2 and Apparent Diffusion Coefficient (ADC)) with pixel-level annotations for the prostate peripheral zone and the transition zone, which are critical for diagnosing and treating prostate cancer. We also do bi-class segmentation for our experiments, all the labeled regions will be mapped to be prostates.}
\item{\textbf{CHAOS}: The CHAOS (Combined Healthy Abdominal Organ Segmentation) dataset is a benchmark dataset designed to support the development and evaluation of abdominal organs. The dataset includes CT and MRI scans, providing a comprehensive set of images that capture different aspects of abdominal organ structures. Our US benchmark needs the multimodal images to be aligned, so we only use the MRI scans under T1-DUAL and T2-SPIR sequences, involving $20$ 3D annotated volumes for organ segmentation.}
\end{itemize}

\section{Experiments}
\begin{figure*}[t]
\centering
\includegraphics[width=1.0\textwidth]{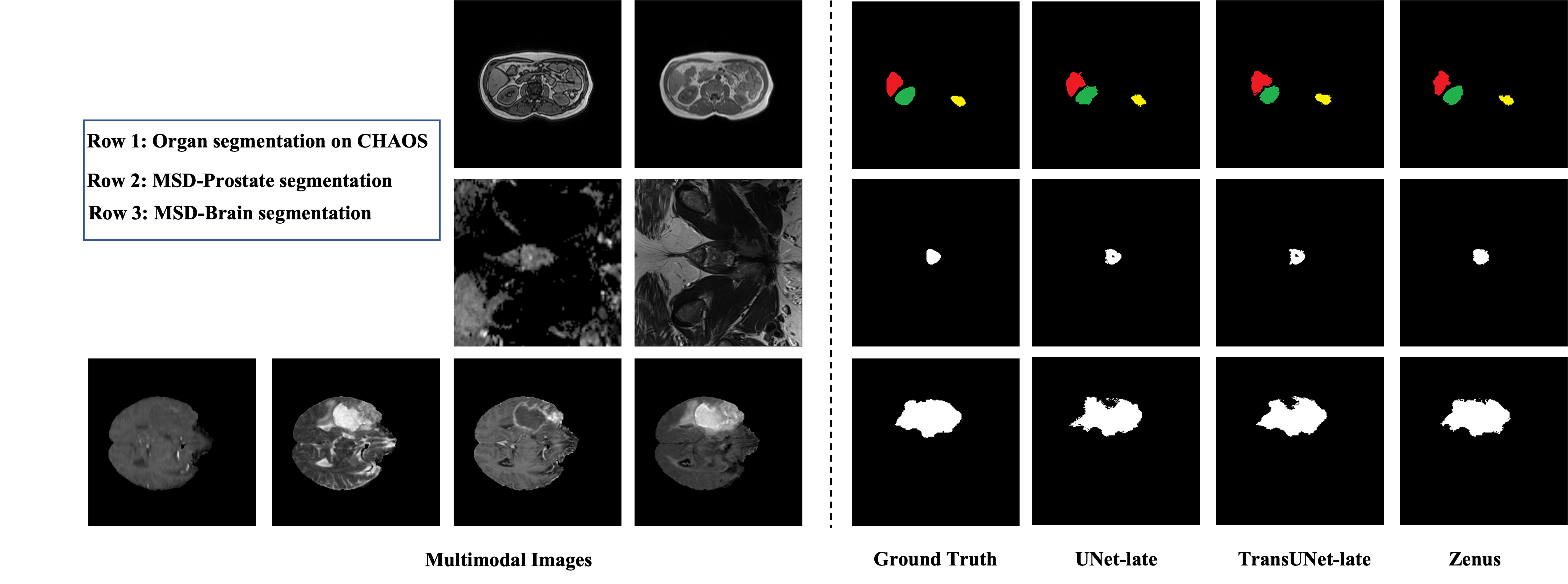}
\caption{Visualization of the multi-class organ segmentation results, bi-class prostate segmentation results, and bi-class brain tumor segmentation results.} 
\label{fig:vis_all}
\end{figure*}

\subsection{Implementation Details and Experimental Setup}
\textbf{Implementation Details.}
We first convert 3D MRI and CT scans into 2D image slices. 
Then, the image slices are resized to $1024 \times 1024$ using nearest interpolation and then adjusted to $256 \times 256$ for instruction generation and model evaluation. 
The training process spans $300$ epochs with an early stopping mechanism activated if the training loss is not reduced for $75$ consecutive epochs. 
The Adam optimizer and synchronized batch normalization are utilized, with a batch size of $10$ and a $l_2$ weight decay of $5e^{-4}$
The initial learning rate are set as $1e-3$ and decayed by $(1-\frac{current\_epoch}{max\_epoch})^{0.9}$. 
Experiments are deployed on 4 $times$ NVIDIA RTX A6000 GPUs. 
We use the commonly employed Dice loss and the BCE loss as our object function.\\

\begin{table*}[t]
\centering
\caption{The first ablation study about the instruction generation module of our proposed methods on three datasets with the late fusion strategy. Blip2 and the QFormer will be employed before the LLM when they are applicable, LoRA module will be added after the LLM when it is applicable in our experiments. The best results are shown in bolded font.}
\setlength{\tabcolsep}{1.5mm}
{
\begin{tabular}{lcccccc}
\hline
\multirow{2}*{Instruction generation module} & \multicolumn{2}{c}{CHAOS} & \multicolumn{2}{c}{MSD-Prostate} & \multicolumn{2}{c}{MSD-Brain}\\ 
 & \multicolumn{1}{c}{DSC$\uparrow$}  &  mIoU$\uparrow$ & \multicolumn{1}{c}{DSC$\uparrow$} & mIoU$\uparrow$ & \multicolumn{1}{c}{DSC$\uparrow$} & mIoU$\uparrow$ \\ \hline 
{Blip2(w/ QFormer)} & \multicolumn{1}{c}{65.15} & {59.06} & \multicolumn{1}{c}{57.56} & {50.95} & \multicolumn{1}{c}{66.70} & {61.48} \\ 
{Blip2(w/ QFormer)+LoRA} & \multicolumn{1}{c}{61.08} & {55.29} & \multicolumn{1}{c}{67.19} & {62.29} & \multicolumn{1}{c}{63.09} & {58.76}\\ \hline
{MedCLIP+QFormer} & \multicolumn{1}{c}{75.59} & {70.14} & \multicolumn{1}{c}{68.29} & {63.29} & \multicolumn{1}{c}{76.43} & {71.70} \\ 
{MedCLIP+QFormer+LoRA} & \multicolumn{1}{c}{72.90} & {67.90} & \multicolumn{1}{c}{67.09} & {65.34} & \multicolumn{1}{c}{74.49} & {70.91}\\ 
{MedCLIP+MLP} & \multicolumn{1}{c}{\textbf{85.80}} & {\textbf{84.19}} & \multicolumn{1}{c}{\textbf{71.09}} & {\textbf{68.36}} & \multicolumn{1}{c}{\textbf{83.67}} & {\textbf{80.86}}\\\hline
\end{tabular}}
\label{table:ab1}
\end{table*}

\begin{table*}[t]
\centering
\caption{The second ablation study about the used pre-trained model of our proposed methods on CHAOS datasets. The best results are shown in bolded font. $LLM_{proj}$ represents the linear projection layer between the VLM encoder and LLM model, LLM for setting under different pre-trained datasets, $F_{p_{enc}}$ regard to the instruction encoder for segmentation prompt and the $Prompt_{backbone}$ as the backbone it based on.}
\setlength{\tabcolsep}{1.5mm}
{
\begin{tabular}{ccccccccccccc}
\hline
\multirow{2}*{Network} & {\multirow{2}*{$F_{enc}$}} &\multirow{2}*{$LLM_{proj}$} &\multirow{2}*{LLM} &\multirow{2}*{$F_{p_{enc}}$} &\multirow{2}*{$Prompt_{backbone}$} & \multicolumn{2}{c}{CHAOS}\\ 
            & & & &  &     & \multicolumn{1}{c}{DSC$\uparrow$} & mIoU$\uparrow$ \\ \hline
\multirow{8}*{Zeus} & {SAM} & $\times$ & Vic-Ori & CLIP & ViT &  \multicolumn{1}{c}{66.34} & {59.81}\\ 
                    & {SAM} & $\times$ & Vic-Ori & MedCLIP & ViT &  \multicolumn{1}{c}{66.21} & {60.96} \\
                    & {SAM} & $\times$ & Vic-Rad & MedCLIP & ViT &  \multicolumn{1}{c}{73.15} & {75.93} \\
                    & {SAM} & $\surd$ & Vic-Ori & CLIP & ViT &  \multicolumn{1}{c}{63.64} & {55.87}\\ 
                    & {SAM} & $\surd$ & Vic-Ori & MedCLIP & ViT &  \multicolumn{1}{c}{68.33} & {61.05} \\
                    & {SAM} & $\surd$ & Vic-Rad & CLIP & ViT &  \multicolumn{1}{c}{78.22} & {64.20} \\
                    & {SAM} & $\surd$ & Vic-Rad & MedCLIP & ResNet &\multicolumn{1}{c}{81.74} & {78.26} \\
                    & {SAM} & $\surd$ & Vic-Rad & MedCLIP & ViT & \multicolumn{1}{c}{83.67} & {80.86} \\
                    & {MedSAM} & {$\surd$} & {Vic-Rad} & {MedCLIP} & {ViT} & \multicolumn{1}{c}{{\textbf{85.33}}} & {{\textbf{82.07}}} \\ \hline
\end{tabular}}
\label{table:ab2}
\end{table*}

\noindent\textbf{Experimental Setup.} 
For a fair comparison, we used pre-trained ResNet and Vision Transformer (ViT) models for every baseline we may use and fine-tuned all the modules without any frozen of the compared baselines, and the fusion strategies are similar to \cite{guo2019deep}, ensuring they are on par with the configuration of our framework. 
Our benchmark and the baselines are validated under three distinct multimodal data fusion strategies: (1) early fusion, combining multimodal input images at the channel dimension, (2) hybrid fusion, integrating image feature maps in the latent space, and (3) late fusion, merging separate masks predicted for each image modality.
To evaluate segmentation accuracy, we utilize the Dice Similarity Coefficient (DSC) and the mean Intersection over Union (mIOU) metrics. These measurements provide insights into the precision and overlap between the predicted segmentation outputs and the ground truths. 
{We divide the framework into two main components: instruction generation and mask prediction. In the instruction generation component, both the LLM and VLM are frozen. As a result, this part does not require fine-tuning and leverages the zero-shot capabilities of pre-trained multimodal models without the need for backward optimization. Unlike \cite{ramesh2021zero,kirillov2023segment}, whose goal is to design models for pre-training and then use the pre-trained weights for zero-shot evaluation, our approach focuses on utilizing existing pre-trained models directly.}

\subsection{Comparative Experiments}
\begin{figure*}[t]
\centering
\includegraphics[width=1.0\textwidth]{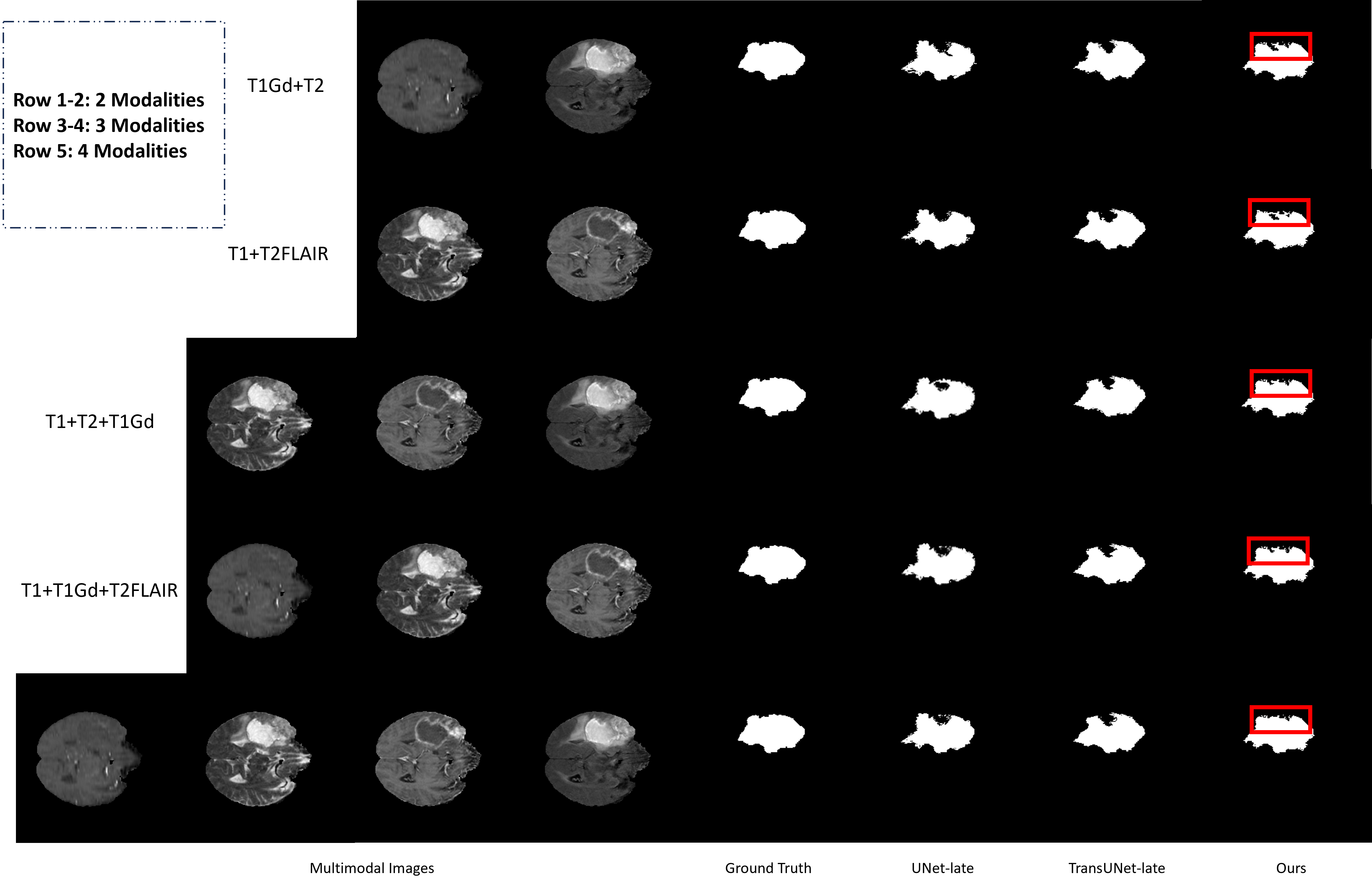}
\caption{Visualization of the bi-class prostate segmentation results, bi-class brain tumor segmentation results, and multi-class organ segmentation results.} 
\label{fig:brain}
\end{figure*}

The main segmentation results as well as the memory of parameters of comparative experiments are shown in Table \ref{table:com}.
It shows that our proposed framework achieves the best DSC across all three datasets. 
In general, the late fusion strategy and the middle fusion strategy almost perform better than the early fusion strategies, which fit intuitive thinking for more parameters and could learn better features.
It also shows that our proposed framework is more efficient compared to the baselines with the smallest trainable network parameter size. The parameter size in the late fusion setting is even smaller than that in the early fusion setting for other baselines. 
We visualize the segmentation results of the $3$ segmentation tasks in Fig. \ref{fig:brain}, we only visualize the baseline result by UNet and TransUNet cause they are the most stable framework under different fusion strategies and stand the most of the second best result compared to other baseline methods. 
All these results demonstrate the superiority of our proposed method. 

\subsection{Ablation Study}
We conducted two comprehensive ablation studies across all three datasets to assess the necessity and significance of each component in our framework. 

The first ablation study explores the instruction generation module shown in Table \ref{table:ab1}, some previously influential LLM-based vision-language models use different alignment strategies for feeding the vision embeddings into LLMs. Blip-v2 \cite{li2023blip} uses a Q-Former module after the vision backbone and uses a question-answer strategy to do alignment by this additional module. And the LISA \cite{lai2023lisa} also followed such settings and made great performance. Blip-v2 cannot fit our framework for it is trained from general data. Even though we added the QFormer module and did not freeze it after the MedCLIP vision encoder, the results were also worse than when we used an MLP-based projection layer which is much better than the original Bilp with the QFormer module. We potentially think that the alignment attributes affect the results when it was aligned with the vision backbone at the training phase, it couldn't help a lot in a zero-shot cross-modality task. LoRA \cite{hu2021lora} is another effective module for processing LLM-related works in text-only tasks, which can also make the segmentation task in \cite{lai2023lisa} be better. However, things even worse when injected into our framework, which is similar when adding LoRA after SAM \cite{kirillov2023segment}. We think a potential reason is that fine-tuning impairs the generalization ability of the used LLMs or LVLMs, especially in the vision-centric tasks cause the generalization gap in vision tasks is much larger than it is in text-only tasks.

For the second ablation study on the CHAOS dataset, we focus on the module alignment in a cross-modality task. Given the dimension discrepancy between the output of the VLM encoder and the input of the LLM, we implemented a projection layer for knowledge transfer and alignment, as suggested by previous works \cite{thawkar2023xraygpt,zhu2023minigpt}. 
The importance of training this layer when adapting the pre-train model from the captioning task to our segmentation task can be demonstrated by comparing the results of the third and last rows in Table \ref{table:ab2}.
Furthermore, we utilize Vicuna as our LLM and CLIP as our instruction prompt encoder, initially trained on broad datasets with minimal medical knowledge. 
The importance of additional fine-tuning on Vicuna and CLIP with a small medical dataset can be demonstrated by comparing the fifth and last rows in Table \ref{table:ab2}. 
The impact of the instruction prompt encoder's knowledge is shown in the comparison of the sixth row to the last row. 
Finally, the choice of backbone for the instruction prompt encoder is significant. As observed in the last two rows, the Vision Transformer (ViT) backbone showcases superior information extraction capability over the ResNet-based backbone. 
In conclusion, comprehension and reasoning abilities concerning medical domain knowledge are crucial, emphasizing the importance of knowledge transfer between cross-modal models. 

\subsection{Deep analysis for Union Segmentation}
\begin{table}[t]
\centering
\caption{The deep analysis experiments for Union Segmentation on CHAOS datasets. The best results are shown in bolded font.}
\setlength{\tabcolsep}{4mm}{
\begin{tabular}{cccccc}
\hline
\multicolumn{4}{c}{Modality} & \multicolumn{2}{c}{CHAOS}\\ 
T1& T1-Gd& T2& T2-FLAIR& \multicolumn{1}{c}{DSC$\uparrow$} & mIoU$\uparrow$ \\ \hline
$\times$ & $\checkmark$ & $\checkmark$ & $\times$ & \multicolumn{1}{c}{80.14} & {78.76}\\
$\checkmark$ & $\times$ & $\times$ & $\checkmark$ &  \multicolumn{1}{c}{80.73} & {79.48} \\
$\checkmark$ & $\checkmark$ & $\checkmark$ & $\times$ &  \multicolumn{1}{c}{81.66} & {79.84} \\
$\checkmark$ & $\checkmark$ & $\times$ & $\checkmark$ & \multicolumn{1}{c}{82.88} & {78.19}\\ 
$\checkmark$ & $\checkmark$ & $\checkmark$ & $\checkmark$ & \multicolumn{1}{c}{\textbf{83.67}} & {\textbf{80.86}} \\\hline
\end{tabular}}
\label{table:us}
\end{table}
We use part of the four modalities in the CHAOS dataset to explore our proposed US benchmark further to validate the performance and compare in Table \ref{table:us}. Besides, the visualization results are in Fig \ref{fig:brain} According to the results, more modalities under the US benchmark could increasingly improve the segmentation accuracy.

\section{{Discussion and Limitations}}
{We acknowledge several limitations in our work. LViT \cite{li2023lvit} demonstrated the effectiveness of task-specific multimodal annotations for segmentation tasks. However, we did not compare the performance of the zero-shot instructions generated by LLM with the task-specific text labels provided by human experts. Bridging the gap between human expertise and large generative models is an important avenue for future research, particularly in the development of advanced visual prompting methods.
Additionally, unlike LLaVA \cite{liu2024visual}, which focuses on language-centric tasks, we did not conduct projection layer-only experiments. Our focus is on a language-to-vision-to-mask pipeline, an image-centric task that cannot be directly addressed by a language model. Developing an LVLM model capable of handling image-centric tasks with fewer trained parameters remains one of our key future goals.
Furthermore, our current fusion strategies are relatively simple and require a significant number of trained parameters. While a late fusion strategy could leverage more parameters and potentially enhance model performance, it may introduce additional computational overhead. In future work, we aim to optimize the fusion module to develop more efficient and effective methods.}

\section{Conclusion}
In this study, we introduce a new benchmark, union segmentation for imitating real-world radiology diagnosis, obtaining different multimodal medical images subject to one object. 
Our new framework (Zeus) uses various pre-trained models. Specifically, we combine LVLM and LLM for zero-shot text instruction generation, leveraging their analytical and reasoning capabilities. Meanwhile, we opted for a lightweight mask decoder module capable of accommodating both image embedding and paired instruction prompts to enhance the effectiveness of mask prediction.
Our Zeus model is assessed through rigorous comparison experiments against influential baselines and ablation studies. The comprehensive results demonstrate the superiority of our novel framework.

\bibliography{sn-bibliography}

\end{document}